\documentclass[10pt,twocolumn,letterpaper]{article}

\usepackage{cvpr}
\usepackage{times}
\usepackage{epsfig}
\usepackage{graphicx}
\usepackage{amsmath}
\usepackage{amssymb}
\usepackage{enumitem}
\usepackage{multirow}
\usepackage{xcolor}
\usepackage{adjustbox}
\newcommand{\widthscale}{0.125}
 
\usepackage[toc,page]{appendix}
\usepackage{caption}
\usepackage{comment}
\usepackage{subcaption}

\usepackage[breaklinks=true,bookmarks=false]{hyperref}

\cvprfinalcopy 

\def\cvprPaperID{****} 

\begin{document}

\title{Zooming Slow-Mo: Fast and Accurate One-Stage Space-Time Video Super-Resolution}


\author{Xiaoyu Xiang$^{\ast}$\\
Purdue University\\
{\tt\small xiang43@purdue.edu}
\and
Yapeng Tian$^{\ast}$\\
University of Rochester\\
{\tt\small yapengtian@rochester.edu}
\and
Yulun Zhang\\
Northeastern University\\
{\tt\small yulun100@gmail.com}
\and
Yun Fu\\
Northeastern University\\
{\tt\small yunfu@ece.neu.edu}
\and
Jan P. Allebach$^{\dagger}$\\
Purdue University\\
{\tt\small allebach@ecn.purdue.edu}
\and
Chenliang Xu$^{\dagger}$\\
University of Rochester\\
{\tt\small chenliang.xu@rochester.edu}
}

\twocolumn[{%
\renewcommand\twocolumn[1][]{#1}%
\maketitle
\begin{center}
    \centering
    \includegraphics[width=\textwidth]{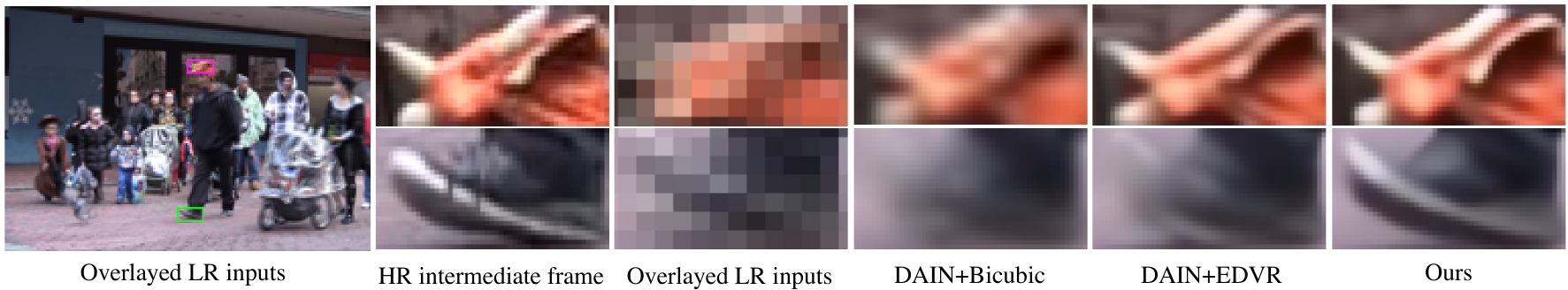}
    \captionof{figure}{\textbf{Example of space-time video super-resolution.} We propose a one-stage space-time video super-resolution (STVSR) network to directly predict high frame rate (HFR) and high-resolution (HR) frames from the corresponding low-resolution (LR) and low frame rate (LFR) frames without explicitly interpolating intermediate LR frames. A HR intermediate frame t and its neighboring low-resolution frames: $t-1$ and $t+1$ as an overlayed image are shown.  Compare to a state-of-the-art two-stage method: DAIN~\cite{bao2019depth}$+$EDVR~\cite{wang2019edvr} on the HR intermediate frame $t$, our method is more capable of handling visual motions and therefore restores more accurate image structures and sharper edges. In addition, our network is more than $3$ times faster on inference speed with a $4$ times smaller model size than the DAIN$+$EDVR.}
\end{center}
}]

\begin{abstract}
In this paper,{\let\thefootnote\relax\footnote{{$^{*}$Equal contribution; $^{\dagger}$\text{Equal advising}.}}} we explore the space-time video super-resolution task, which aims to generate a high-resolution (HR) slow-motion video from a low frame rate (LFR), low-resolution (LR) video. A simple solution is to split it into two sub-tasks: video frame interpolation (VFI) and video super-resolution (VSR). However, temporal interpolation and spatial super-resolution are intra-related in this task. Two-stage methods cannot fully take advantage of the natural property. In addition, state-of-the-art VFI or VSR networks require a large frame-synthesis or reconstruction module for predicting high-quality video frames, which makes the two-stage methods have large model sizes and thus be time-consuming. To overcome the problems, we propose a one-stage space-time video super-resolution framework, which directly synthesizes an HR slow-motion video from an LFR, LR video. Rather than synthesizing missing LR video frames as VFI networks do, we firstly temporally interpolate LR frame features in missing LR video frames capturing local temporal contexts by the proposed feature temporal interpolation network. Then, we propose a deformable ConvLSTM to align and aggregate temporal information simultaneously for better leveraging global temporal contexts. Finally, a deep reconstruction network is adopted to predict HR slow-motion video frames. Extensive experiments on benchmark datasets demonstrate that the proposed method not only achieves better quantitative and qualitative performance but also is more than three times faster than recent two-stage state-of-the-art methods, \eg, DAIN+EDVR and DAIN+RBPN. 
\end{abstract}\

\section{Introduction}

Space-Time Video Super-Resolution (STVSR)~\cite{shechtman2005space} aims to automatically generate a photo-realistic video sequence with a high space-time resolution from a low-resolution and low frame rate input video. Since HR slow-motion videos are more visually appealing containing fine image details and clear motion dynamics, they are desired in rich applications, such as film making and high-definition television.



To tackle the problem, most existing works in previous literatures~\cite{shechtman2005space,mudenagudi2010space,takeda2010spatiotemporal,shahar2011space,faramarzi2012space,li2015space} usually adopt hand-crafted regularization and make strong assumptions. For example, space-time directional smoothness prior is adopted in ~\cite{shechtman2005space}, and \cite{mudenagudi2010space} assumes that there is no significant change in illumination for the static pixels. However, these strong constraints make the methods have limited capacity in modeling various and diverse space-time visual patterns. Besides, the optimization for these methods is usually computationally expensive (\eg, $\sim 1$ hour for $60$ frames in~\cite{mudenagudi2010space}).

In recent years, deep convolutional neural networks have shown promising efficiency and effectiveness in various video restoration tasks, such as video frame interpolation (VFI)~\cite{niklaus2017adavonv}, video super-resolution (VSR)~\cite{caballero2017real}, and video deblurring~\cite{su2017deep}. To design an STVSR network, one straightforward way is by directly combining a video frame interpolation method (\eg, SepConv~\cite{niklaus2017adsconv}, ToFlow~\cite{xue2019video}, DAIN~\cite{bao2019depth} \etc) and a video super-resolution method (\eg, DUF~\cite{jo2018deep}, RBPN~\cite{haris2019recurrent}, EDVR~\cite{wang2019edvr} \etc) in a two-stage manner. It firstly interpolates missing intermediate LR video frames with VFI and then reconstructs all HR frames with VSR. However, temporal interpolation and spatial super-resolution in STVSR are intra-related. The two-stage methods splitting them into two individual procedures cannot make full use of this natural property. 
In addition, to predict high-quality video frames, both state-of-the-art VFI and VSR networks require a big frame reconstruction network. Therefore, the composed two-stage STVSR model will contain a large number of parameters and is computationally expensive. 

To alleviate the above issues, we propose a unified one-stage STVSR framework to learn temporal interpolation and spatial super-resolution simultaneously. We propose to adaptively learn a deformable feature interpolation function for temporally interpolating intermediate LR frame features rather than synthesizing pixel-wise LR frames as in two-stage methods. The learnable offsets in the interpolation function can aggregate useful local temporal contexts and help the temporal interpolation handle complex visual motions. In addition, we introduce a new deformable ConvLSTM model to effectively leverage global contexts with simultaneous temporal alignment and aggregation. HR video frames can be reconstructed from the aggregated LR features with a deep SR reconstruction network. To this end, the one-stage network can learn end-to-end to map an LR, LFR video sequence to its HR, HFR space in a sequence-to-sequence manner. Experimental results show that the proposed one-stage STVSR framework outperforms state-of-the-art two-stage methods even with much fewer parameters. An example is illustrated in Figure \textcolor{red}{1}. 



The contributions of this paper are three-fold: (1) We propose a one-stage space-time super-resolution network that can address temporal interpolation and spatial SR simultaneously in a unified framework. Our one-stage method is more effective than two-stage methods taking advantage of the intra-relatedness between the two sub-problems. It is also computationally more efficient since only one frame reconstruction network is required rather than two large networks as in state-of-the-art two-stage approaches. (2) We propose a frame feature temporal interpolation network leveraging local temporal contexts based on deformable sampling for intermediate LR frames. We devise a novel deformable ConvLSTM to explicitly enhance temporal alignment capacity and exploit global temporal contexts for handling large motions in videos. (3) Our one-stage method achieves state-of-the-art STVSR performance on both Vid4 \cite{liu2011bayesian} and Vimeo~\cite{xue2019video}. It is $3$ times faster than the two-stage network: DAIN~\cite{bao2019depth} + EDVR~\cite{wang2019edvr} while having a nearly $4\times$ reduction in model size. {\textcolor{magenta}{The source code is released in \href{https://github.com/Mukosame/Zooming-Slow-Mo-CVPR-2020}{https://github.com/Mukosame/Zooming-SlowMo-CVPR-2020}.}}

\begin{figure*}
    \centering
    \includegraphics[width=0.9\textwidth]{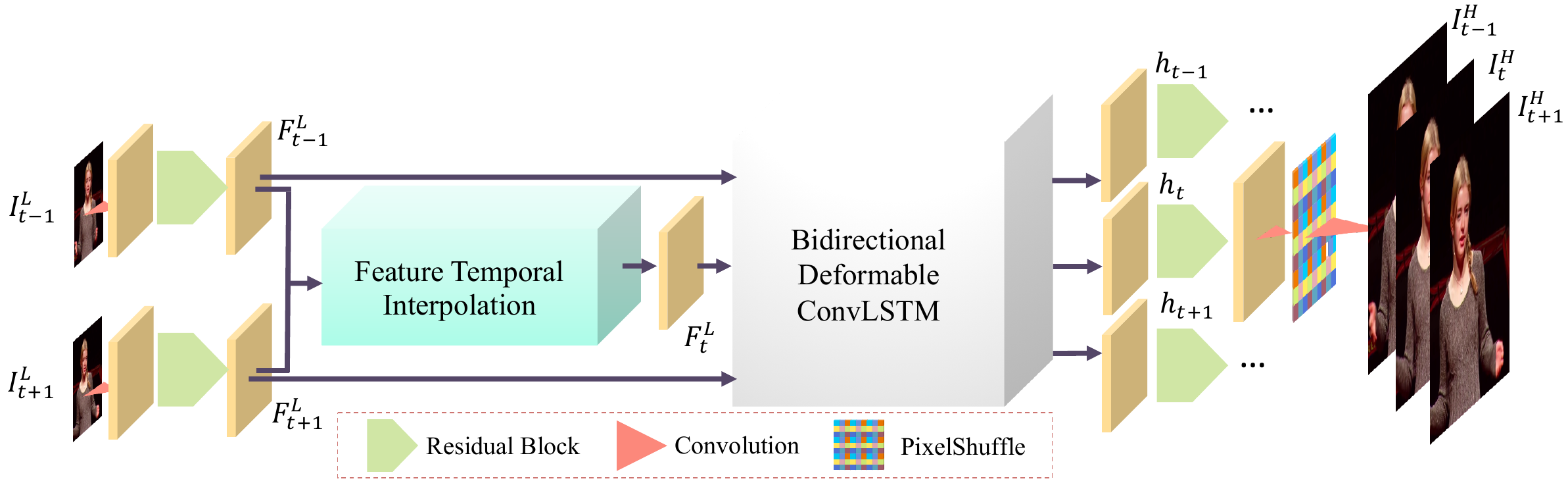}
    \caption{Overview of our one-stage STVSR framework. It directly reconstructs consecutive HR video frames without synthesizing LR intermediate frames $I_t^L$. Feature temporal interpolation and bidirectional deformable ConvLSTM are utilized to leverage local and global temporal contexts for better exploiting temporal information and handling large motions. Note that we only show two input LR frames from a long sequence in this figure for a better illustration.}
    \label{fig:structure}
    \vspace{-3mm}
\end{figure*}

\section{Related Work}
In this section, we discuss works on three related topics: video frame interpolation (VFI), video super-resolution (VSR), and space-time video super-resolution (STVSR).

\begin{description}[style=unboxed,leftmargin=0cm]

\item[Video Frame Interpolation]
The target of video frame interpolation is to synthesize non-existent intermediate frames in between the original frames. 
Meyer~\etal~\cite{meyer2015phase} introduced a phase-based frame interpolation method, which generates intermediate frames through per-pixel phase modification. Long \etal \cite{long2016learning} predicted intermediate frames directly with an encoder-decoder CNN. Niklaus~\etal~\cite{niklaus2017adavonv,niklaus2017adsconv} regarded the frame interpolation as a local convolution over the two input frames and used a CNN to
learn a spatially-adaptive convolution kernel for each pixel for high-quality frame synthesis. To explicitly handle motions, there are also many flow-based video interpolation approaches~\cite{jiang2018super,liu2017video,niklaus2018context, bao2019memc, bao2019depth}. These methods usually have inherent issues with inaccuracies and missing information from optical flow results. 
In our one-stage STVSR framework, rather than synthesizing the intermediate LR frames as current VFI methods do, we interpolate features from two neighboring LR frames to directly synthesize LR feature maps for missing frames without requiring explicit supervision.

\item[Video Super-Resolution]
Video super-resolution aims to reconstruct an HR video frame from the corresponding LR frame (reference frame) and its neighboring LR frames (supporting frames). One key problem for VSR is how to temporally align the LR supporting frames with the reference frame. Several VSR methods~\cite{caballero2017real,tao2017detail,sajjadi2018frame,wang2018learning,xue2019video} use optical flow for explicit temporal alignment, which first estimates motions between the reference frame and each supporting frame with optical flow and then warps the supporting frame using the predicted motion map. Recently, RBPN proposes to incorporate the single image and multi-frame SR for VSR in which flow maps are directly concatenated with LR video frames. However, it is difficult to obtain accurate flow; and flow warping also introduces artifacts into the aligned frames. To avoid this problem, DUF~\cite{jo2018deep} with dynamic filters and TDAN~\cite{tian2018tdan} with deformable alignment were proposed for implicit temporal alignment without motion estimation. EDVR~\cite{wang2019edvr} extends the deformable alignment in TDAN by exploring multiscale information. However, most of the above methods are many-to-one architectures, and they need to process a batch of LR frames to predict only one HR frame, which makes the methods computationally inefficient. 
Recurrent neural networks, such as convolutional LSTMs~\cite{xingjian2015convolutional} (ConvLSTM), can ease sequence-to-sequence (S2S) learning; and they are adopted in VSR methods~\cite{lim2017deep,huang2017video} for leveraging temporal information. However, without explicit temporal alignment, the RNN-based VSR networks have limited capability in handling large and complex motions within videos. 
To achieve efficient yet effective modeling, unlike existing methods, we propose a novel ConvLSTM structure embedded with an explicit state updating cell for space-time video super-resolution.

Rather than simply combining a VFI network and a VSR network to solve STVSR, we propose a more efficient and effective one-stage framework that simultaneously learns temporal feature interpolation and spatial SR without accessing to LR intermediate frames as supervision.
\item[Space-Time Video Super-Resolution]
Shechtman \etal~\cite{shechtman2002increasing} firstly proposed to extend SR to the space-time domain. Since pixels are missing in LR frames and even several entire LR frames are unavailable, STVSR is a highly ill-posed inverse problem. To increase video resolution both in time and space, \cite{shechtman2002increasing} combines information from multiple video sequences of dynamic scenes obtained at sub-pixel and sub-frame misalignments with a directional space-time smoothness regularization to constrain the ill-posed problem. Mudenagudi~\cite{mudenagudi2010space} posed STVSR as a reconstruction
problem using the Maximum a posteriori-Markov Random Field~\cite{geman1984stochastic} with graph-cuts~\cite{boykov2001fast} as the solver. Takeda~\etal~\cite{takeda2010spatiotemporal} exploited local orientation and local motion to steer spatio-temporal regression kernels. Shahar~\etal~\cite{shahar2011space} proposed to exploit a space-time patch recurrence prior within natural videos for STVSR. However, these methods have limited capacity to model rich and complex space-time visual patterns, and the optimization for these methods is usually computationally expensive. To address these issues, we propose a one-stage network to directly learn the mapping between partial LR observations and HR video frames and to achieve fast and accurate STVSR. 
\end{description}

\section{Space-Time Video Super-Resolution}
\label{STVSR}

Given an LR, LFR video sequence: $\mathcal{I}^{L}=\{I_{2t-1}^L\}_{t=1}^{n+1}$, our goal is to generate the corresponding high-resolution slow-motion video sequence: $\mathcal{I}^{H}=\{I_{t}^H\}_{t=1}^{2n+1}$. To intermediate HR frames $\{I_{2t}^H\}_{t=1}^{n}$, there are no corresponding LR counterparts in the input sequence. To fast and accurately increase resolution in both space and time domains, we propose a one-stage space-time super-resolution framework: Zooming Slow-Mo as illustrated in Figure~\ref{fig:structure}. The framework mainly consists of four parts: \textit{feature extractor}, \textit{frame feature temporal interpolation module}, \textit{deformable ConvLSTM}, and \textit{HR frame reconstructor}.

We first use a feature extractor with a convolutional layer and $k_1$ residual blocks to extract feature maps: $\{F_{2t-1}^L\}_{t=1}^{n+1}$ from input video frames. Taking the feature maps as input, we then synthesize the LR feature maps: $\{F_{2t}^L\}_{t=1}^{n}$ of intermediate frames with the proposed frame feature interpolation module. Furthermore, to better leverage temporal information, we use a deformable ConvLSTM to process the consecutive feature maps: $\{F_{t}^L\}_{t=1}^{2n+1}$. Unlike vanilla ConvLSTM, the proposed deformable ConvLSTM can simultaneously perform temporal alignment and aggregation. Finally, we reconstruct the HR slow-mo video sequence from the aggregated feature maps. 


\begin{figure}
    \centering
    \includegraphics[width=1.0\columnwidth]{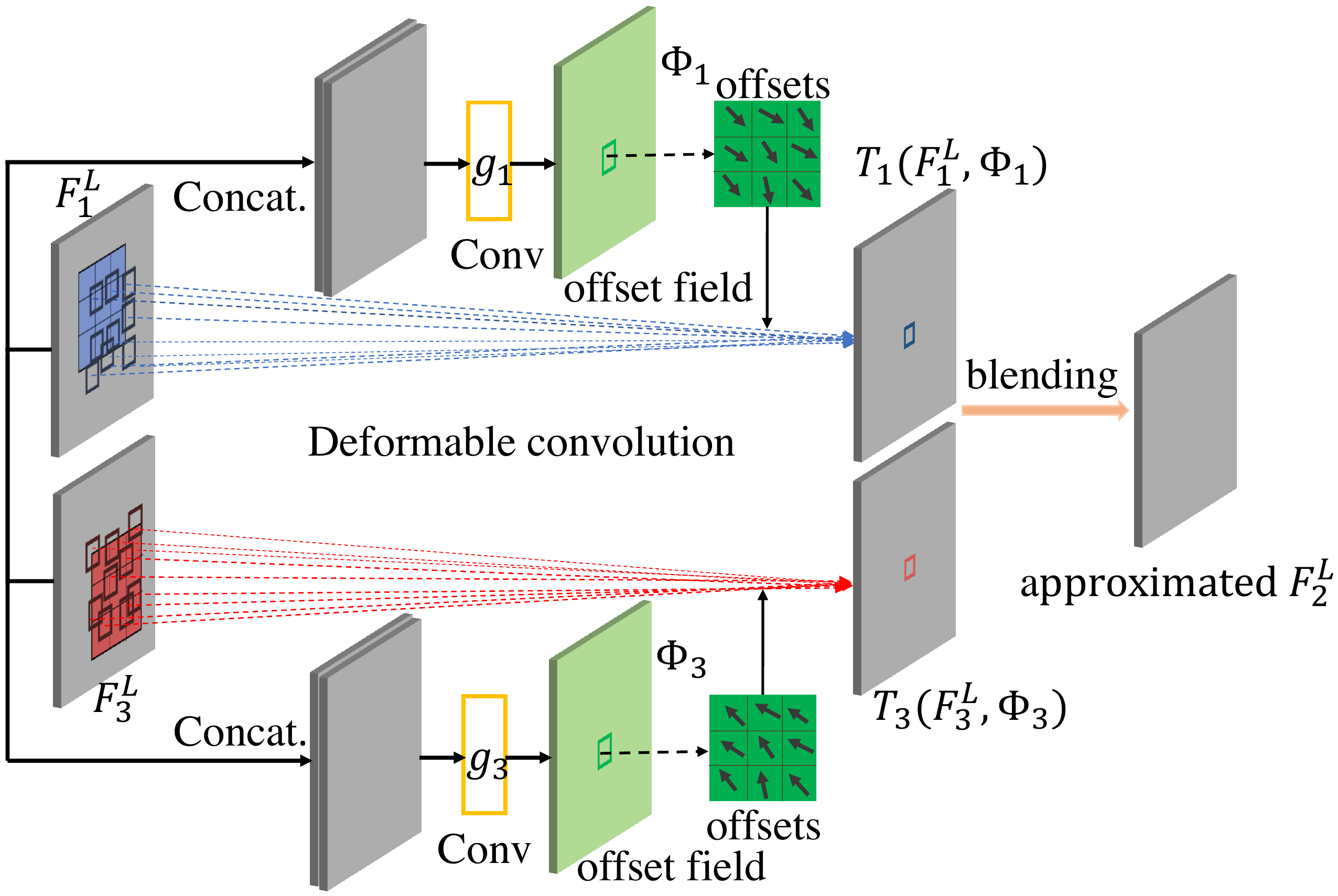}
    \caption{Frame feature temporal interpolation based on deformable sampling. Since approximated $F_2^L$ will be used to predict the corresponding HR frame, it will implicitly enforce the learnable offsets to capture accurate local temporal contexts and be motion-aware. }
    \label{fig:interp_module}
    \vspace{-5mm}
\end{figure}

\subsection{Frame Feature Temporal Interpolation}
\label{dcn}

Given extracted feature maps: $F_{1}^L$ and $F_{3}^L$ from input LR video frames: $I_{1}^L$ and $I_{3}^L$, we want to synthesize the feature map $F_{2}^L$ corresponding to the missing intermediate LR frame $I_{2}^L$. Traditional video frame interpolation networks usually perform temporal interpolation on pixel-wise video frames, which will lead to a two-stage STVSR design. Unlike previous methods, we propose to learn a feature temporal interpolation function $f(\cdot)$ to directly synthesize the intermediate feature map $F_{2}^L$ (see Fig.~\ref{fig:interp_module}). A general form of the interpolation function can be formulated as:
\begin{equation}
    F_{2}^L = f(F_{1}^L, F_{3}^L) = H(T_{1}(F_{1}^L, \Phi_1), T_{3}(F_{3}^L, \Phi_3))
    \enspace,
\end{equation}
where $T_1(\cdot)$ and $T_3(\cdot)$ are two sampling functions and $\Phi_1$ and $\Phi_3$ are the corresponding sampling parameters; $H(\cdot)$ is a blending function to aggregate sampled features.

For generating accurate $F_{2}^L$, the $T_1(\cdot)$ should capture forward motion information between $F_{1}^L$ and $F_{2}^L$, and the $T_3(\cdot)$ should capture backward motion information between $F_{3}^L$ and $F_{2}^L$. However, the $F_{2}^L$ is not available for computing forward and backward motion information in this task. 

To alleviate this problem, we use motion information between $F_{1}^L$ and $F_{3}^L$ to approximate forward and backward motion information. Inspired by recent deformable alignment in~\cite{tian2018tdan} for VSR, we propose to use deformable sampling functions to implicitly capture motion information for frame feature temporal interpolation. With exploring rich local temporal contexts by deformable convolutions in sampling functions, our feature temporal interpolation can even handle very large motions in videos.

The two sampling functions share the same network design but have different weights. For simplicity, we use the $T_1(\cdot)$ as an example. It takes LR frame feature maps $F_{1}^L$ and $F_{3}^L$ as input to predict an offset for sampling the $F_{1}^L$:
\begin{equation}
    \Delta p_{1} = g_1([F_{1}^L, F_{3}^L])
    \enspace,
\end{equation}
where $\Delta p_1$ is a learnable offset and also refers to the sampling parameter: $\Phi_1$; $g_1$ denotes a general function of several convolution layers; $[ ,]$ denotes the channel-wise concatenation. With the learned offset, the sampling function can be performed with a deformable convolution~\cite{dai2017deformable, zhu2019deformable}: 
\begin{equation}
  T_1(F_1^L, \Phi_1) = DConv(F_{1}^L, \Delta p_{1})
  \enspace.  
\end{equation}
 Similarly, we can learn an offset $\Delta p_{3} = g_3([F_{3}^L, F_{1}^L])$ as the sampling parameter: $\Phi_3$ and then obtain sampled features $T_3(F_3^L, \Phi_3)$ with a deformable convolution.

To blend the two sampled features, we use a simple linear blending function $H(\cdot)$:
\begin{equation}
   F_{2}^{L}  = \alpha * T_{1}(F_{1}^L, \Phi_1) + \beta * T_{3}(F_{3}^L, \Phi_3)
   \enspace,
\end{equation}
where $\alpha$ and $\beta$ are two learnable $1\times1$ convolution kernels and $*$ is a convolution operator. Since the synthesized LR feature map $F_2^L$ will be used to predict the intermediate HR frame $I_2^H$, it will enforce the synthesized LR feature map to be close to the real intermediate LR feature map. Therefore, the two offsets $\Delta p_{1}$ and $\Delta p_{3}$ will implicitly learn to capture the forward and backward motion information, respectively.

Applying the designed deformable temporal interpolation function to $\{F_{2t-1}^L\}_{t=1}^{n+1}$, we can obtain intermediate frame feature maps $\{F_{2t}^L\}_{t=1}^{n}$.


\subsection{Deformable ConvLSTM}
\label{dconvlstm}

Now we have consecutive frame feature maps: $\{F_{t}^L\}_{t=1}^{2n+1}$ for generating the corresponding HR video frames, which will be a sequence-to-sequence mapping. It has been proved in previous video restoration tasks~\cite{xue2019video, tao2017detail, wang2019edvr} that temporal information is vital. Therefore, rather than reconstructing HR frames from the corresponding individual feature maps, we aggregate temporal contexts from neighboring frames. ConvLSTM \cite{xingjian2015convolutional} is a popular 2D sequence data modeling method and we can adopt it to perform temporal aggregation. At the time step $t$, the ConvLSTM updates hidden state $h_{t}$ and cell state $c_{t}$ with:
\begin{align}
h_t, c_t = ConvLSTM(h_{t-1},c_{t-1},{F}_t^{L})
\enspace.
\end{align}
From its state updating mechanism \cite{xingjian2015convolutional}, we can learn that the ConvLSTM can only implicitly capture motions between previous states: $h_{t-1}$ and $c_{t-1}$ and the current input feature map with small convolution receptive fields. Therefore, ConvLSTM has limited ability to handle large motions in natural videos. If a video has large motions, there will be a severe temporal mismatch between previous states and $F_t^L$. Then, $h_{t-1}$ and $c_{t-1}$ will propagate mismatched ``noisy'' content rather than useful global temporal contexts into $h_t$. Consequently, the reconstructed HR frame $I_t^H$ from $h_t$ will suffer from annoying artifacts.  
 
\begin{figure}
    \centering
    \includegraphics[width=1.0\columnwidth]{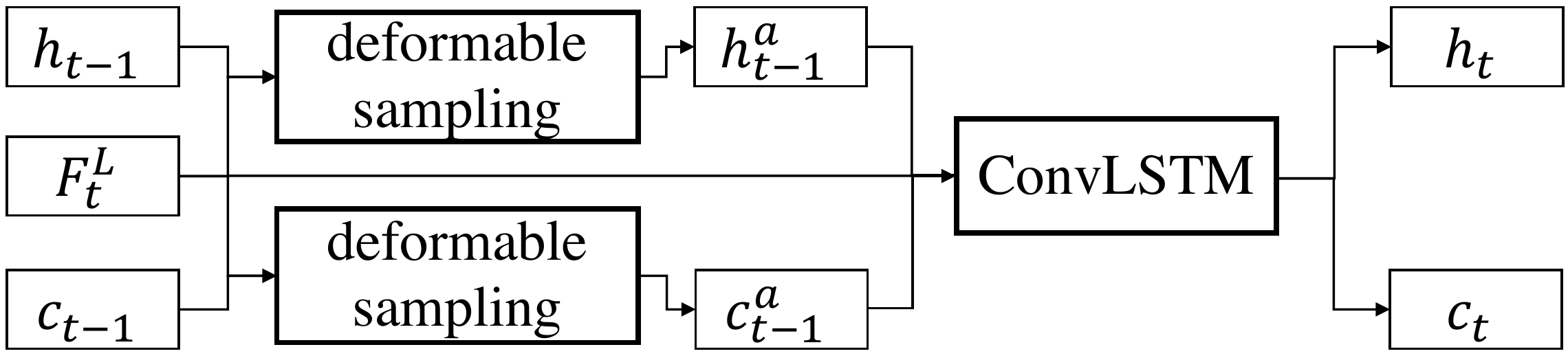}
    \caption{Deformable ConvLSTM for better exploiting global temporal contexts and handling fast motion videos. At time step $t$, we introduce state updating cells to learn deformable sampling to adaptively align hidden state $h_{t-1}$ and cell state $c_{t-1}$ with current input feature map: $F_t^L$.}
    \label{fig:conv_lstm}
\vspace{-5mm}
\end{figure}

To tackle the large motion problem and effectively exploit global temporal contexts, we explicitly embed a state-updating cell with deformable alignment into ConvLSTM (see Fig.~\ref{fig:conv_lstm}): 
 \begin{equation}
\begin{split}
& \Delta p^h_{t} = g^h([h_{t-1},F_t^L ]) \enspace,\\
& \Delta p^c_{t} = g^c([c_{t-1},F_t^L ]) \enspace,\\
& h^a_{t-1} = DConv(h_{t-1}, \Delta p^h_{t}) \enspace,\\
& c^a_{t-1} = DConv(c_{t-1}, \Delta p^c_{t}) \enspace,\\
&h_t, c_t = ConvLSTM(h^a_{t-1}, c^a_{t-1},F_t^L) \enspace,
\end{split}
\end{equation}
where $g^h$ and $g^c$ are general functions of several convolution layers, $\Delta p^h_{t}$ and $\Delta p^c_{t}$ are predicted offsets, and $h^a_{t-1}$ and $c^a_{t-1}$ are aligned hidden and cell states, respectively.
Compared with vanilla ConvLSTM, we explicitly enforce the hidden state $h_{t-1}$ and cell state $c_{t-1}$ to align with the current input feature map $F_t^L $ in our deformable ConvLSTM, which makes it more capable of handling motions in videos. Besides, to fully explore temporal information, we use the Deformable ConvLSTM in a bidirectional manner~\cite{schuster1997bidirectional}. We feed temporally reversed feature maps into the same Deformable ConvLSTM and concatenate hidden states from forward pass and backward pass as the final hidden state $h_t$\footnote{We use $h_t$ to denote final hidden state, but it will refer to a concatenated hidden state in the Bidirectional Deformable ConvLSTM.} for HR frame reconstruction. 

\subsection{Frame Reconstruction}
To reconstruct HR video frames, we use a temporally shared synthesis network, which takes individual hidden state $h_t$ as input and outputs the corresponding HR frame. It has $k_2$ stacked residual blocks~\cite{lim2017enhanced} for learning deep features  and utilizes a sub-pixel upscaling module with PixelShuffle as in \cite{shi2016real} to reconstruct HR frames $\{I_{t}^H\}_{t=1}^{2n+1}$. To optimize our network, we use a reconstruction loss function:
\begin{equation}
l_{rec} = \sqrt{||I^{GT}_{t} - I_{t}^H ||^2 + \epsilon^2}
\enspace,
\end{equation}
where $I^{GT}_{t}$ refers to the $t$-th ground-truth HR video frame, Charbonnier penalty function~\cite{lai2017deep} is used as the loss term, and $\epsilon$ is empirically set to $1\times10^{-3}$. 
Since the space and time SR problems are intra-related in STVSR, our model is end-to-end trainable and can simultaneously learn this spatio-temporal interpolation with only supervision from HR video frames. 

\subsection{Implementation Details}

In our implementation, $k_1 = 5$ and $k_2 = 40$ residual blocks are used in feature extraction and HR frame reconstruction modules, respectively. We randomly crop a sequence of down-sampled image patches with the size of $32\times32$ and take out the odd-indexed 4 frames as LFR and LR inputs, and the corresponding consecutive 7-frame sequence of $4\times$\footnote{Considering recent state-of-the-art methods (\eg, EDVR~\cite{wang2019edvr} and RBPN~\cite{haris2019recurrent}) use only 4 as the upscaling factor, we adopt the same practice.} size as supervision. Besides, we perform data augmentation by randomly rotating $90^{\circ}$, $180^{\circ}$ and $270^{\circ}$, and horizontal-flipping. We adopt a Pyramid, Cascading and Deformable (PCD) structure in~\cite{wang2019edvr} to employ deformable alignment and apply Adam \cite{kingma2014adam} optimizer, where we decay the learning rate with a cosine annealing for each batch \cite{loshchilov2016sgdr} from $4e-4$ to $1e-7$. The batch size is set to be 24 and trained on $2$ Nvidia Titan XP GPUs.

\section{Experiments and Analysis}
\subsection{Experimental Setup}

\begin{figure*}[htbp]
	\scriptsize
	\centering
	\begin{tabular}{cc}
        \begin{adjustbox}{valign=t}
			\begin{tabular}{c}
				\includegraphics[width=0.255\textwidth]{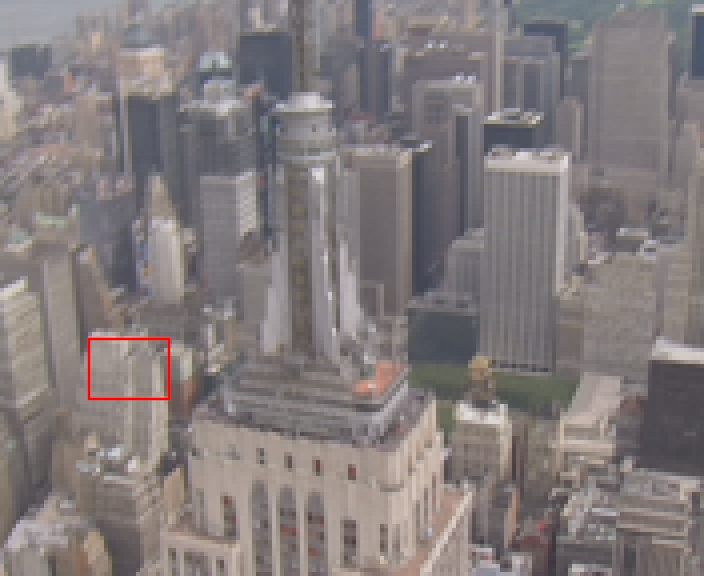}
				\\
				Overlayed LR frames
			
			\end{tabular}
		\end{adjustbox}
		\hspace{-4.3mm}
		\begin{adjustbox}{valign=t}
			\begin{tabular}{ccccc}
				\includegraphics[width=\widthscale \textwidth]{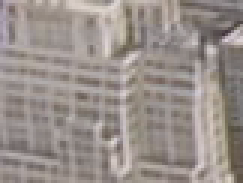} \hspace{-4mm} &
				\includegraphics[width=\widthscale \textwidth]{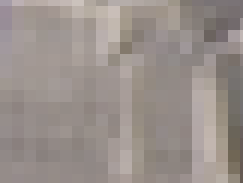} \hspace{-4mm} &
				\includegraphics[width=\widthscale \textwidth]{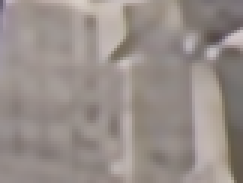} \hspace{-4mm} &
				\includegraphics[width=\widthscale \textwidth]{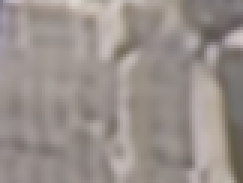}\hspace{-3.5mm} &
				\includegraphics[width=\widthscale \textwidth]{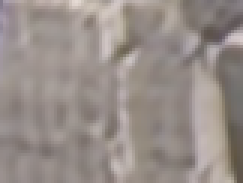}
				\\
				HR \hspace{-4mm} &
    			Overlayed LR \hspace{-4mm} &
				SepConv+RCAN \hspace{-4mm} &
				SepConv+RBPN\hspace{-4mm} &
				SepConv+EDVR

				\\
				\includegraphics[width=\widthscale \textwidth]{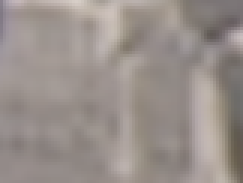} \hspace{-4mm} &
				\includegraphics[width=\widthscale \textwidth]{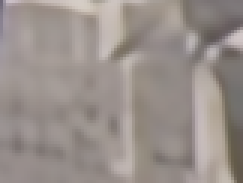} \hspace{-4mm} &
				\includegraphics[width=\widthscale \textwidth]{Figs/fig2/city/city_004_dain_rbpn.png} \hspace{-4mm} &
				\includegraphics[width=\widthscale \textwidth]{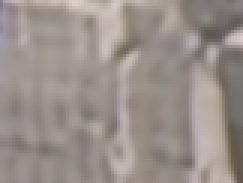}\hspace{-3.5mm} &
				\includegraphics[width=\widthscale \textwidth]{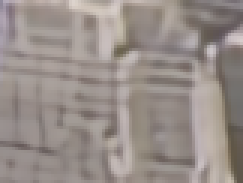}
				\\
				DAIN+Bicubic \hspace{-4mm} &
				DAIN+RCAN \hspace{-4mm} &
				DAIN+RPBN\hspace{-4mm} &
				DAIN+EDVR\hspace{-4mm} &
				\textbf{Ours}
				\\
			\end{tabular}
			\end{adjustbox}

         \\
    \begin{adjustbox}{valign=t}
			\begin{tabular}{c}
				\includegraphics[width=0.255\textwidth]{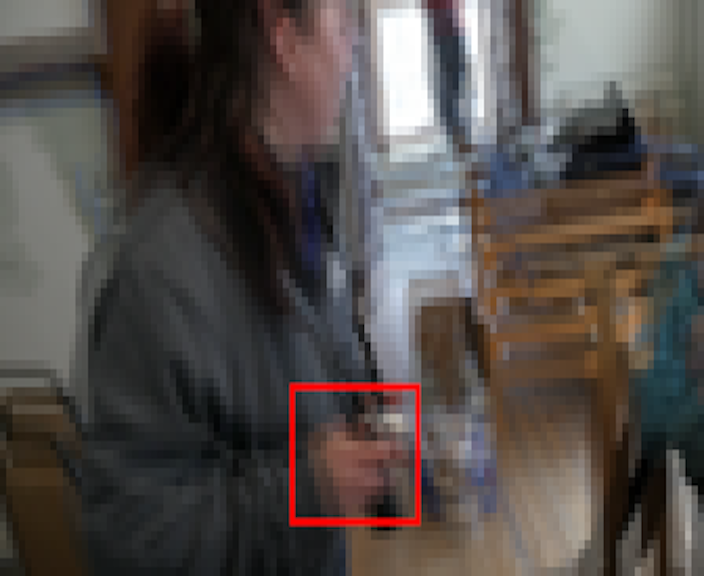}
				\\
				Overlayed LR frames
			
			\end{tabular}
		\end{adjustbox}
		\hspace{-4.3mm}
		\begin{adjustbox}{valign=t}
			\begin{tabular}{ccccc}
				\includegraphics[width=\widthscale \textwidth]{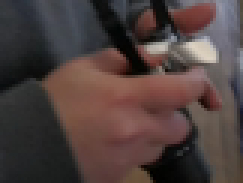} \hspace{-4mm} &
				\includegraphics[width=\widthscale \textwidth]{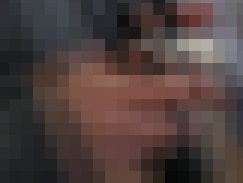} \hspace{-4mm} &
				\includegraphics[width=\widthscale \textwidth]{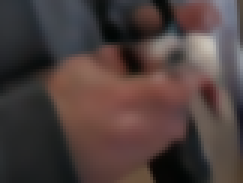} \hspace{-4mm} &
				\includegraphics[width=\widthscale \textwidth]{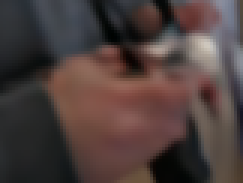}\hspace{-3.5mm} &
				\includegraphics[width=\widthscale \textwidth]{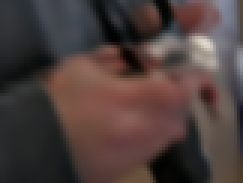}
				\\
			    HR \hspace{-4mm} &
    			Overlayed LR \hspace{-4mm} &
				SepConv+RCAN \hspace{-4mm} &
				SepConv+RBPN\hspace{-4mm} &
				SepConv+EDVR

				\\
				\includegraphics[width=\widthscale \textwidth]{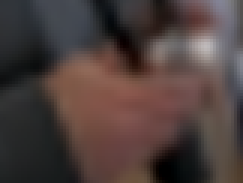} \hspace{-4mm} &
				\includegraphics[width=\widthscale \textwidth]{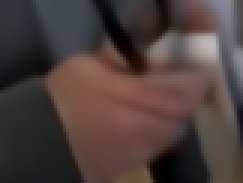} \hspace{-4mm} &
				\includegraphics[width=\widthscale \textwidth]{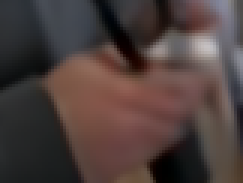} \hspace{-4mm} &
				\includegraphics[width=\widthscale \textwidth]{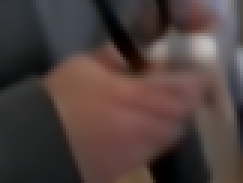}\hspace{-3.5mm} &
				\includegraphics[width=\widthscale \textwidth]{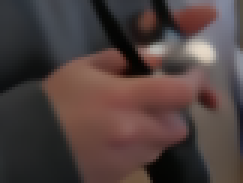}
				\\
			    DAIN+Bicubic \hspace{-4mm} &
				DAIN+RCAN \hspace{-4mm} &
				DAIN+RPBN\hspace{-4mm} &
				DAIN+EDVR\hspace{-4mm} &
				\textbf{Ours}
				\\
			\end{tabular}
			\end{adjustbox}
					      \\
    \begin{adjustbox}{valign=t}
			\begin{tabular}{c}
				\includegraphics[width=0.255\textwidth]{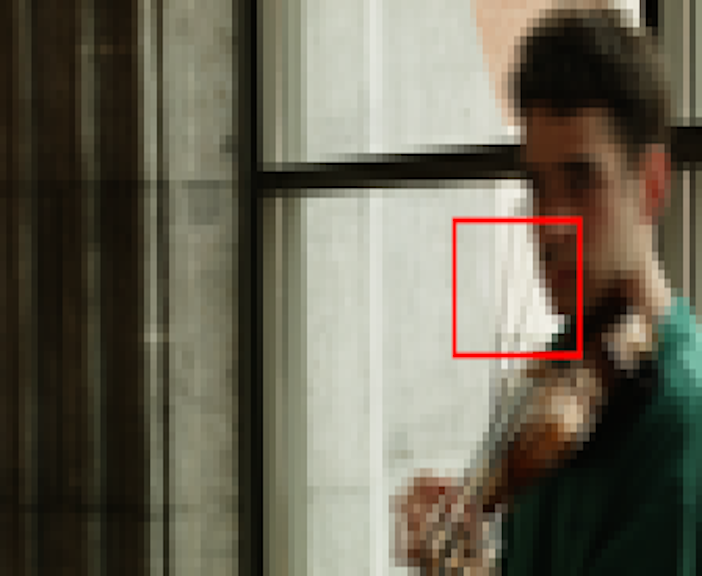}
				\\
				Overlayed LR frames
			
			\end{tabular}
		\end{adjustbox}
		\hspace{-4.3mm}
		\begin{adjustbox}{valign=t}
			\begin{tabular}{ccccc}
				\includegraphics[width=\widthscale \textwidth]{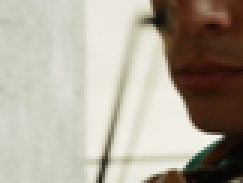} \hspace{-4mm} &
				\includegraphics[width=\widthscale \textwidth]{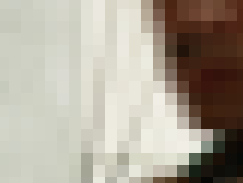} \hspace{-4mm} &
				\includegraphics[width=\widthscale \textwidth]{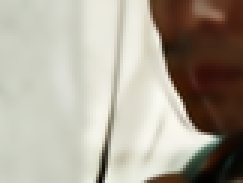} \hspace{-4mm} &
				\includegraphics[width=\widthscale \textwidth]{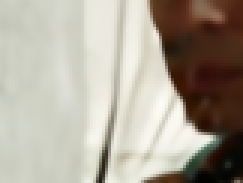}\hspace{-3.5mm} &
				\includegraphics[width=\widthscale \textwidth]{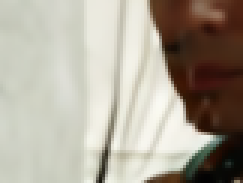}
				\\
			    HR \hspace{-4mm} &
    			Overlayed LR \hspace{-4mm} &
				SepConv+RCAN \hspace{-4mm} &
				SepConv+RBPN\hspace{-4mm} &
				SepConv+EDVR

				\\
				\includegraphics[width=\widthscale \textwidth]{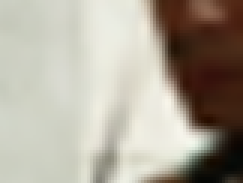} \hspace{-4mm} &
				\includegraphics[width=\widthscale \textwidth]{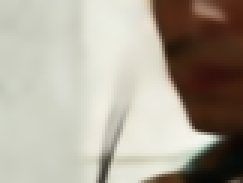} \hspace{-4mm} &
				\includegraphics[width=\widthscale \textwidth]{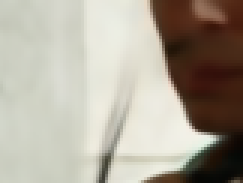} \hspace{-4mm} &
				\includegraphics[width=\widthscale \textwidth]{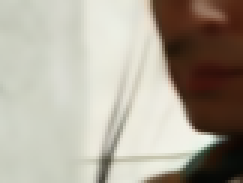}\hspace{-3.5mm} &
				\includegraphics[width=\widthscale \textwidth]{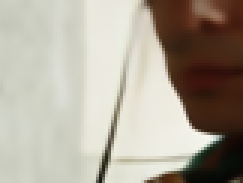}
				\\
			    DAIN+Bicubic \hspace{-4mm} &
				DAIN+RCAN \hspace{-4mm} &
				DAIN+RPBN\hspace{-4mm} &
				DAIN+EDVR\hspace{-4mm} &
				\textbf{Ours}
				\\
			\end{tabular}
			\end{adjustbox}
			
							      \\
    \begin{adjustbox}{valign=t}
			\begin{tabular}{c}
				\includegraphics[width=0.255\textwidth]{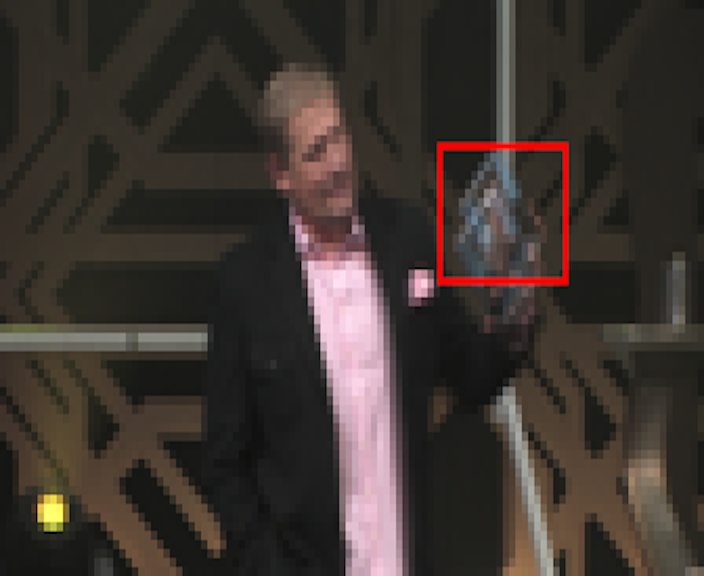}
				\\
				Overlayed LR frames
			
			\end{tabular}
		\end{adjustbox}
		\hspace{-4.3mm}
		\begin{adjustbox}{valign=t}
			\begin{tabular}{ccccc}
				\includegraphics[width=\widthscale \textwidth]{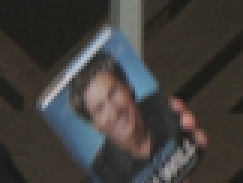} \hspace{-4mm} &
				\includegraphics[width=\widthscale \textwidth]{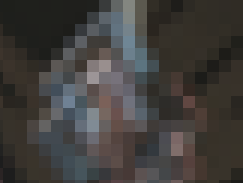} \hspace{-4mm} &
				\includegraphics[width=\widthscale \textwidth]{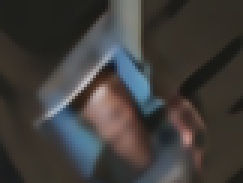} \hspace{-4mm} &
				\includegraphics[width=\widthscale \textwidth]{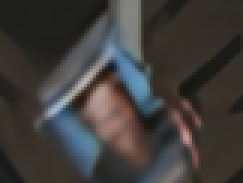}\hspace{-3.5mm} &
				\includegraphics[width=\widthscale \textwidth]{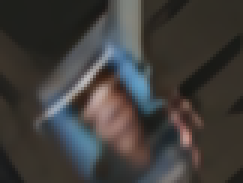}
				\\
			    HR \hspace{-4mm} &
    			Overlayed LR \hspace{-4mm} &
				SepConv+RCAN \hspace{-4mm} &
				SepConv+RBPN\hspace{-4mm} &
				SepConv+EDVR

				\\
				\includegraphics[width=\widthscale \textwidth]{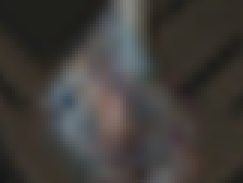} \hspace{-4mm} &
				\includegraphics[width=\widthscale \textwidth]{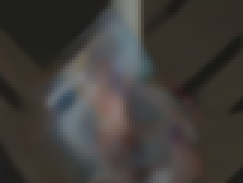} \hspace{-4mm} &
				\includegraphics[width=\widthscale \textwidth]{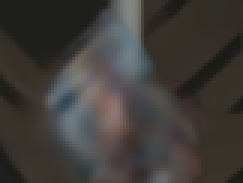} \hspace{-4mm} &
				\includegraphics[width=\widthscale \textwidth]{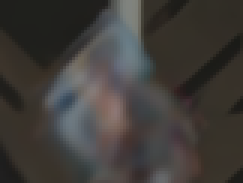}\hspace{-3.5mm} &
				\includegraphics[width=\widthscale \textwidth]{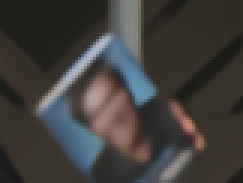}
				\\
			    DAIN+Bicubic \hspace{-4mm} &
				DAIN+RCAN \hspace{-4mm} &
				DAIN+RPBN\hspace{-4mm} &
				DAIN+EDVR\hspace{-4mm} &
				\textbf{Ours}
				\\
			\end{tabular}
			\end{adjustbox}
  
	\end{tabular}
	\caption{Visual comparisons of different methods on video frames from Vid4 and Vimeo datasets. Our one-stage Zooming SlowMo model can reconstruct more visually appealing HR video frames with more accurate image structures and fewer blurring artifacts. }
	\label{fig:stvsrresult}
	\vspace{-5mm}
\end{figure*}

\begin{description}[style=unboxed,leftmargin=0cm]
\item[Datasets]
We use Vimeo-90K as the training set \cite{xue2019video}, including more than 60,000 7-frame training video sequences. The dataset is widely used in previous VFI and VSR works \cite{bao2019memc, bao2019depth, tian2018tdan, haris2019recurrent, wang2019edvr}. Vid4~\cite{liu2011bayesian} and Vimeo testset \cite{xue2019video} are used as the evaluation datasets. To measure the performance of different methods under different motion conditions, we split the Vimeo testset into fast motion, medium motion, and slow motion sets as in ~\cite{haris2019recurrent}, which include $1225$, $4977$ and $1613$ video clips, respectively. We remove $5$ video clips from the original medium motion set and $3$ clips from the slow motion set, which have consecutively all-black background frames that will lead to infinite values on PSNR.
We generate LR frames by bicubic with a downsampling factor 4 and use odd-indexed LR frames as input to predict the corresponding consecutive HR and HFR frames.

\item[Evaluation] Peak Signal-to-Noise Ratio (PSNR) and Structural Similarity Index (SSIM) \cite{wang2004image} are adopted to evaluate STVSR performance of different methods. To measure the efficiency of different networks, we also compare the model sizes and inference time of the entire Vid4 \cite{liu2011bayesian} dataset measured on one Nvidia Titan XP GPU.


\end{description}

\subsection{Comparison to State-of-the-art Methods}

We compare the performance of our one-stage Zooming SlowMo network to two-stage methods composed of state-of-the-art (SOTA) VFI and VSR networks.
Three recent SOTA VFI approaches, SepConv \cite{niklaus2017adsconv}, Super-SloMo\footnote{Since there is no official source code released, we used an unofficial PyTorch implementation from  \hyperlink{https://github.com/avinashpaliwal/Super-SloMo}{https://github.com/avinashpaliwal/Super-SloMo}.}~\cite{jiang2018super}, and DAIN \cite{bao2019depth}, are compared. To achieve STVSR, three SOTA SR models, including single-image SR model, RCAN \cite{zhang2018image}, and two recent VSR models, RBPN \cite{haris2019recurrent} and EDVR  \cite{wang2019edvr}, are used to generate HR frames from both original LR and interpolated LR frames.

\begin{table*}[h]
\caption{Quantitative comparison of our results and two-stage VFI and VSR methods on testsets. The best two results are highlighted in \textcolor{red}{red} and \textcolor{blue}{blue} colors, respectively. The total runtime is measured on the entire Vid4 dataset \cite{liu2011bayesian}. Note that we omit the baseline models with Bicubic when comparing in terms of runtime.}
\resizebox{\textwidth}{!}{
\begin{tabular}{cc|ccccccccccccc}
\hline
\multirow{2}{*}{\begin{tabular}[c]{@{}c@{}}VFI\\   Method\end{tabular}} & \multirow{2}{*}{\begin{tabular}[c]{@{}c@{}}SR\\   Method\end{tabular}} & \multicolumn{2}{c}{Vid4} & \multicolumn{2}{c}{Vimeo-Fast} & \multicolumn{2}{c}{Vimeo-Medium} & \multicolumn{2}{c}{Vimeo-Slow} & \multirow{2}{*}{\begin{tabular}[c]{@{}c@{}}Parameters\\   (Million)\end{tabular}} & \multirow{2}{*}{\begin{tabular}[c]{@{}c@{}}Runtime-VFI\\   (s)\end{tabular}} & \multirow{2}{*}{\begin{tabular}[c]{@{}c@{}}Runtime-SR\\   (s)\end{tabular}} & \multirow{2}{*}{\begin{tabular}[c]{@{}c@{}}Total\\   Runtime (s)\end{tabular}} & \multirow{2}{*}{\begin{tabular}[c]{@{}c@{}}Average\\   Runtime (s/frame)\end{tabular}} \\ 
                                                                        &                                                                        & PSNR       & SSIM        & PSNR          & SSIM           & PSNR           & SSIM            & PSNR          & SSIM           &                                                                                   &                                                                              &                                                                             &                                                                                &                                                                                        \\ \hline \hline
SuperSloMo \cite{jiang2018super}                                                                 & Bicubic                                                                   &   22.84 &   	0.5772 &   	31.88 &   	0.8793 &   	29.94 &   	0.8477 &   	28.37 &   	0.8102 &   	19.8 &   	0.28 &   	- &   	- &   	-                                                                               \\
SuperSloMo \cite{jiang2018super}                                                                & RCAN \cite{zhang2018image}                                                                  & 23.80      & 0.6397      & 34.52         & 0.9076         & 32.50          & 0.8884          & 30.69         & 0.8624         & 19.8+16.0                                                                         & 0.28                                                                         & 68.15                                                                       & 68.43                                                                          & 0.4002                                                                                 \\
SuperSloMo \cite{jiang2018super}                                                                & RBPN \cite{haris2019recurrent}                                                                  & 23.76      & 0.6362      & 34.73         & 0.9108    &      32.79 & 	0.8930	 & 30.48 & 	0.8584 & 	\textcolor{blue}{19.8+12.7} & 	0.28 & 	82.62 & 	82.90 & 	0.4848        \\
SuperSloMo \cite{jiang2018super}                                                                & EDVR \cite{wang2019edvr}                                                                  & 24.40  & 	0.6706	  & 35.05  & 	0.9136  & 	33.85  & 	0.8967  & 	30.99  & 	0.8673  & 	19.8+20.7  & 	0.28  & 	24.65  & 	\textcolor{blue}{24.93}  & 	\textcolor{blue}{0.1458}                                                                        \\ \hline
SepConv \cite{niklaus2017adsconv}                                                                 & Bicubic                                                                   &   23.51 &   	0.6273 &   	32.27 &   	0.8890 &   	30.61 &   	0.8633 &   	29.04 &   	0.8290 &   	21.7 &   	2.24 &   	- &   	- &   	-                                                           \\
SepConv \cite{niklaus2017adsconv}                                                                & RCAN \cite{zhang2018image}                                                                  & 24.92      & 0.7236      & 34.97         & 0.9195         & 33.59          & 0.9125          & 32.13         & 0.8967         & 21.7+16.0                                                                         & 2.24                                                                         & 68.15                                                                       & 70.39                                                                          & 0.4116                                                                                 \\
SepConv \cite{niklaus2017adsconv}                                                                & RBPN \cite{haris2019recurrent}                                                                  & 26.08      & 0.7751      & 35.07         & 0.9238         &       34.09         &      0.9229           & 32.77         & 0.9090         & 21.7+12.7                                                                         & 2.24                                                                    & 82.62                                                                       & 84.86                                                                          & 0.4963                                                                                 \\
SepConv \cite{niklaus2017adsconv}                                                                & EDVR \cite{wang2019edvr}                                                                  & 25.93      & 0.7792      & 35.23         &       0.9252         & 34.22          &     0.9240            & 32.96         &          0.9112      & 21.7+20.7                                                                         & 2.24                                                                         & 24.65                                                                       & 26.89                                                                          & 0.1572                                                                                 \\
\hline
DAIN \cite{bao2019depth}                                                                   & Bicubic                                                                &    23.55  &   	0.6268  &   	32.41  &   	0.8910  &   	30.67  &   	0.8636  &   	29.06	  &   0.8289  &   	24.0	  &   8.23  &   	-  &   	-  &   	-                         \\
DAIN  \cite{bao2019depth}                                                                  & RCAN \cite{zhang2018image}                                                                  & 25.03      & 0.7261      & 35.27         & 0.9242         & 33.82          & 0.9146          & 32.26         & 0.8974         & 24.0+16.0                                                                         & 8.23                                                                         &           68.15                                                                  & 76.38                                                                           & 0.4467                                                                                 \\
DAIN  \cite{bao2019depth}                                                                  & RBPN \cite{haris2019recurrent}                                                                  & 25.96      & 0.7784      & 35.55         & 0.9300         & 34.45          & 0.9262          & 32.92       & 0.9097         & 24.0+12.7                                                                         & 8.23                                                                         &           82.62                                                                  & 90.85                                                                           & 0.5313                                                                                 \\
DAIN \cite{bao2019depth}                                                                   & EDVR \cite{wang2019edvr}                                                                  & \textcolor{blue}{26.12}      & \textcolor{blue}{0.7836}        & \textcolor{blue}{35.81}         &     \textcolor{blue}{0.9323}       & \textcolor{blue}{34.66}          &     \textcolor{blue}{0.9281}            & \textcolor{blue}{33.11}         &      \textcolor{blue}{0.9119}          & 24.0+20.7                                                                         & 8.23                                                                         & 24.65                                                                       & 32.88                                                                          & 0.1923                                                                                 \\ \hline
\multicolumn{2}{c|}{Ours}                                                                                                                         & \textcolor{red}{26.31}      &      \textcolor{red}{0.7976}       & \textcolor{red}{36.81}         &      \textcolor{red}{0.9415}          & \textcolor{red}{35.41}          &      \textcolor{red}{0.9361}           & \textcolor{red}{33.36}         &      \textcolor{red}{0.9138}         & \textcolor{red}{11.10}                                                                             & -                                                                            & -                                                                           & \textcolor{red}{10.36}                                                                          & \textcolor{red}{0.0606}                                                                                
 
\\ \hline
\end{tabular}
}
\label{tab:result}
\vspace{-3mm}
\end{table*}

Quantitative results are shown in Table \ref{tab:result}. From the table, we can learn the following facts: (1) DAIN+EDVR is the best performing two-stage approach among the compared 12 methods; (2) VFI matters, especially for fast motion videos. Although RBPN and EDVR perform much better than RCAN for VSR, however, when equipped with more advanced VFI network DAIN, DAIN+RCAN can achieve comparable or even better performance than SepConv+RBPN and SepConv+EDVR on the Vimeo -Fast set; (3) VSR also matters. For example, with the same VFI network: DAIN, EDVR consistently achieves better STVSR performance than other VSR methods. In addition, we can see that our network outperforms the DAIN+EDVR by $0.19$dB on Vid4, $0.25$dB on Vimeo-Slow, $0.75$dB on Vimeo-Medium, and $1$dB on Vimeo-Fast in terms of PSNR. The significant improvements obtained on videos with fast motions demonstrate that our one-stage network with simultaneously leveraging local and global temporal contexts is more capable of handling diverse spatio-temporal patterns, including challenging large motions in videos than two-stage methods. 

Moreover, we also investigate model sizes and runtime of different networks in Table \ref{tab:result}. For synthesizing high-quality frames, SOTA VFI and VSR networks usually have very large frame reconstruction modules. Thus, the composed two-stage SOTA STVSR networks will contain a huge number of parameters. With only one frame reconstruction module, our one-stage model has much fewer parameters than the SOTA two-stage networks. From Table \ref{tab:result}, we can see that it is more than $4\times$ and $3\times$ smaller than the DAIN+EDVR and DAIN+RBPN, respectively. The small model size makes our network more than $3\times$ faster than the DAIN+EDVR and $8\times$ faster than DAIN+RBPN. Compared to two-stage methods with a fast VFI network: SuperSlowMo, our method is still more than $2\times$ faster. 



Visual results of different methods are illustrated in Figure~\ref{fig:stvsrresult}. We see that our method achieves noticeably visual improvements over other two-stage methods. Clearly, the proposed network can synthesize visually appealing HR video frames with more fine details, more accurate structures, and fewer blurring artifacts even for challenging fast motion video sequences. We also observe that current SOTA VFI methods: SepConv and DAIN fail to handle large motions. Consequently, two-stage networks tend to generate HR frames with severe motion blurs. In our one-stage framework, we simultaneously learn temporal and spatial SR with exploring the natural intra-relatedness. Even with a much smaller model, our network can well address the large motion issue in temporal SR. 

\subsection{Ablation Study}
We have already shown the superiority of our one-stage framework over two-stage networks. To further demonstrate the effectiveness of different modules in our network, we make a comprehensive ablation study. 

\begin{figure}[tb]
\captionsetup[subfigure]{labelformat=empty}
\begin{center}
  \begin{subfigure}[b]{0.42\linewidth}
     \includegraphics[width=\linewidth]{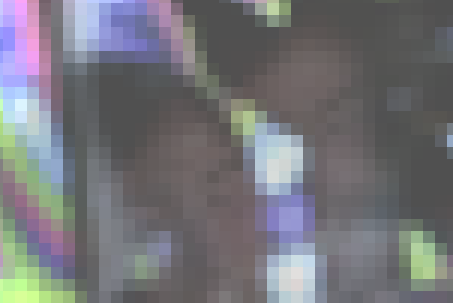}
     \subcaption{Overlayed LR}
  \end{subfigure}
  \begin{subfigure}[b]{0.42\linewidth}
  \includegraphics[width=\linewidth]{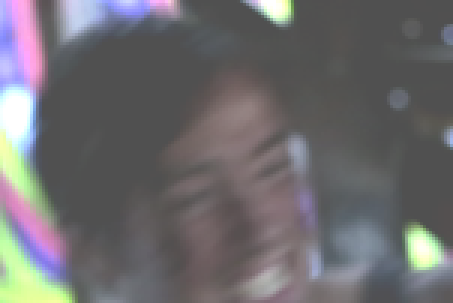}
  \subcaption{HR}
  \end{subfigure}
  
  \begin{subfigure}[b]{0.42\linewidth}
     \includegraphics[width=\linewidth]{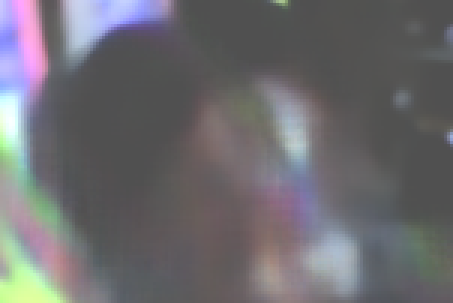}
     \subcaption{w/o DFI@model (a)}
  \end{subfigure}
  \begin{subfigure}[b]{0.42\linewidth}
  \includegraphics[width=\linewidth]{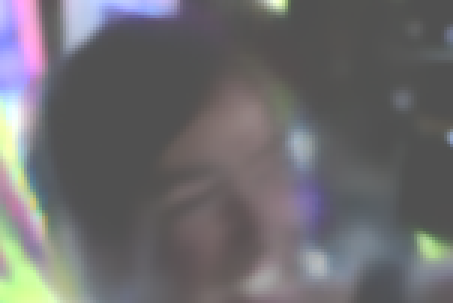}
  \subcaption{w/ DFI@model (b)}
  \end{subfigure}
\end{center}
\vspace{-5mm}
   \caption{Ablation study on feature interpolation. The naive feature interpolation model without deformable sampling will obtain overly smooth results for videos with fast motions. With the proposed deformable feature interpolation (DFI), our model can well exploit local contexts in adjacent frames, thus is more effective in handling large motions.}
 \label{fig:ablation_interp}
\vspace{-4mm}
\end{figure}

\begin{table}
\caption{Ablation study on the proposed modules. Proposed deformable feature interpolation network and deformable ConvLSTM can effectively handle motions and improve STVSR performance, while the vanilla ConvLSTM performs worse when meeting large motions in videos.}
\resizebox{\columnwidth}{!}{
\begin{tabular}{l|cc|ccc}
\hline
Method                                                                        & (a) & (b) & (c)   & (d)     & (e)  \\ \hline \hline
Naive feature interpolation                                                                  &     $\surd$  &          &                  &           &              \\
Deformable feature interpolation (DFI)     &      & $\surd$      & $\surd$      & $\surd$      & $\surd$              \\ \hline
ConvLSTM                                                                      &            &          & $\surd$      &          &                        \\
Deformable ConvLSTM (DConvLSTM)                                                         &           &          &          & $\surd$      &                  \\
Bidirectional DConvLSTM                                                             &      &          &          &          & $\surd$                \\ 
 \hline \hline
Vid4 (slow motion)                                                                    &   25.18        & 25.34 & 25.68 & 26.18 & 26.31          \\
\hline
Vimeo-Fast (fast motion)                                                                     &     34.93 &  35.66 &  35.39 & 36.56 & 36.81\\ 
\hline
\end{tabular}
}
\label{tab:ablation}
\end{table}


\begin{figure*}[tb]
\captionsetup[subfigure]{labelformat=empty}
\begin{center}
  \begin{subfigure}[b]{0.18\linewidth}
     \includegraphics[width=\linewidth]{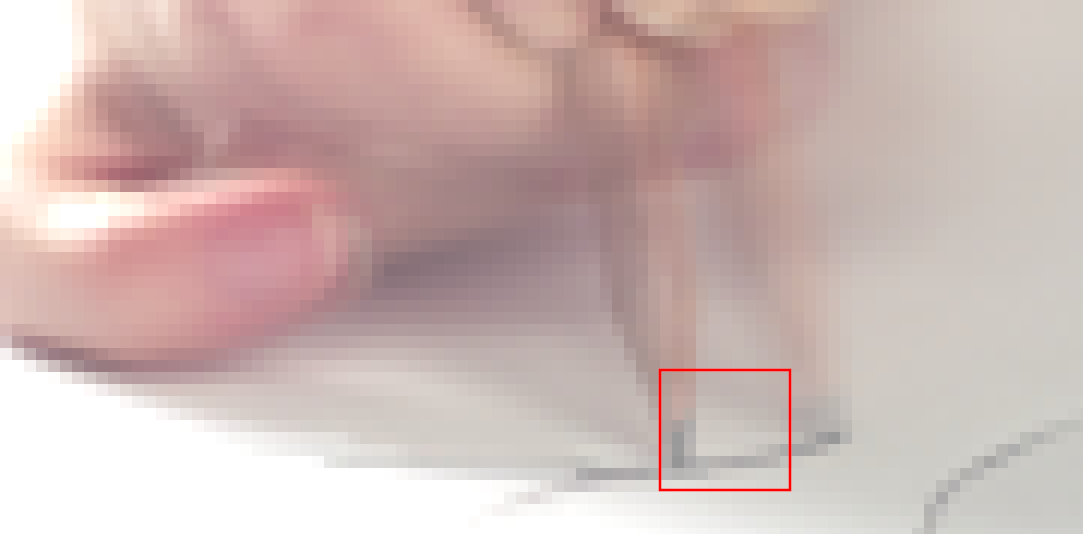}
    
  \end{subfigure}
  \begin{subfigure}[b]{0.18\linewidth}
  \includegraphics[width=\linewidth]{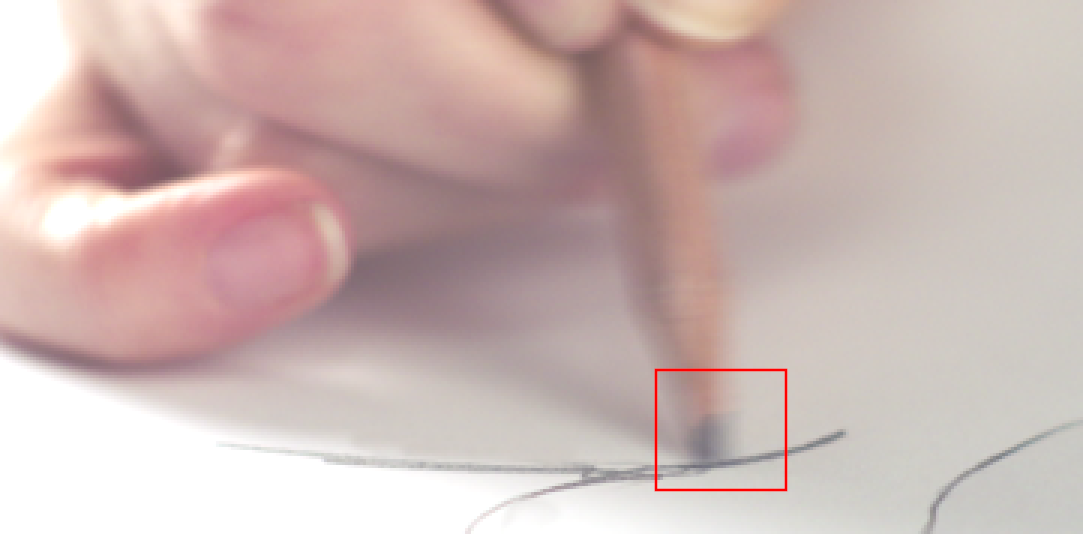}

  \end{subfigure}
  \begin{subfigure}[b]{0.18\linewidth}
     \includegraphics[width=\linewidth]{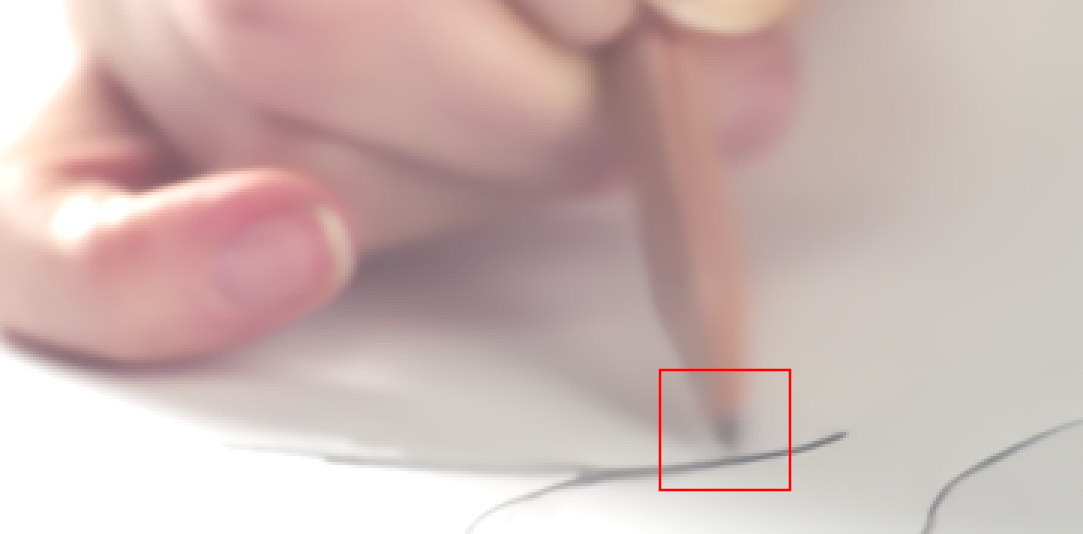}
     
  \end{subfigure}
  \begin{subfigure}[b]{0.18\linewidth}
  \includegraphics[width=\linewidth]{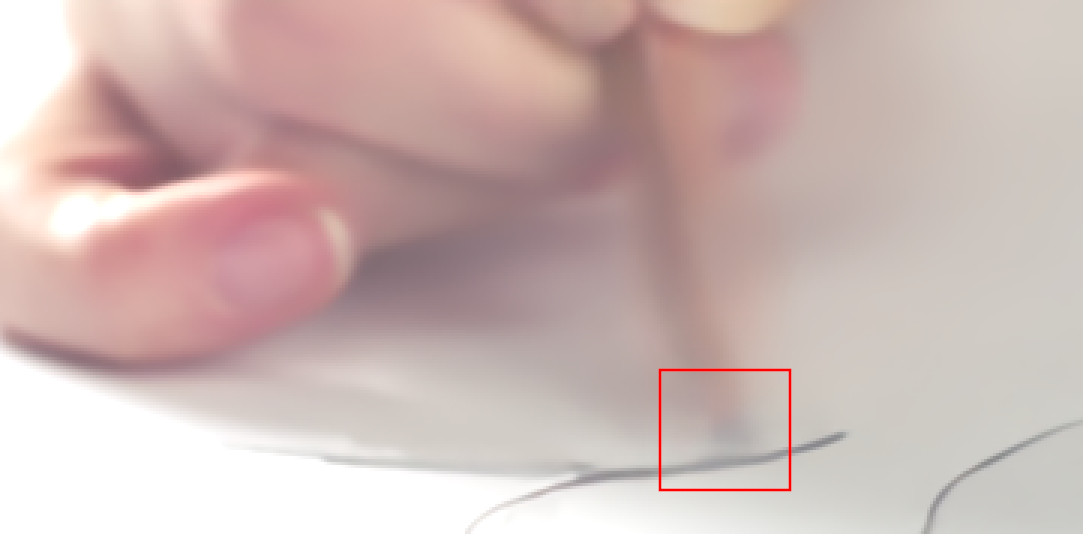}
 
  \end{subfigure}
  \begin{subfigure}[b]{0.18\linewidth}
  \includegraphics[width=\linewidth]{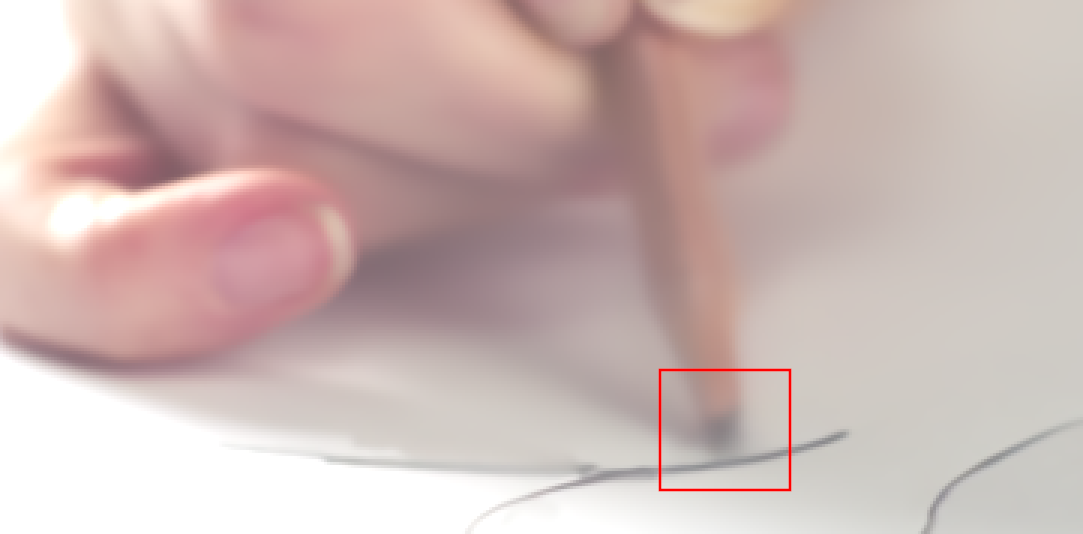}
  \end{subfigure}
  
  \begin{subfigure}[b]{0.18\linewidth}
     \includegraphics[width=\linewidth]{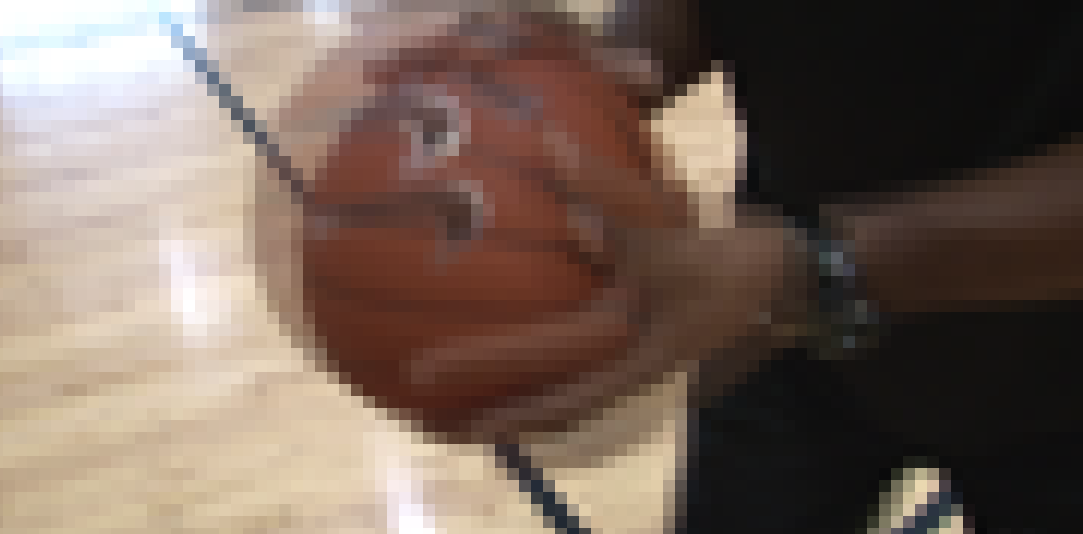}
     \subcaption{Overlayed LR}
  \end{subfigure}
  \begin{subfigure}[b]{0.18\linewidth}
  \includegraphics[width=\linewidth]{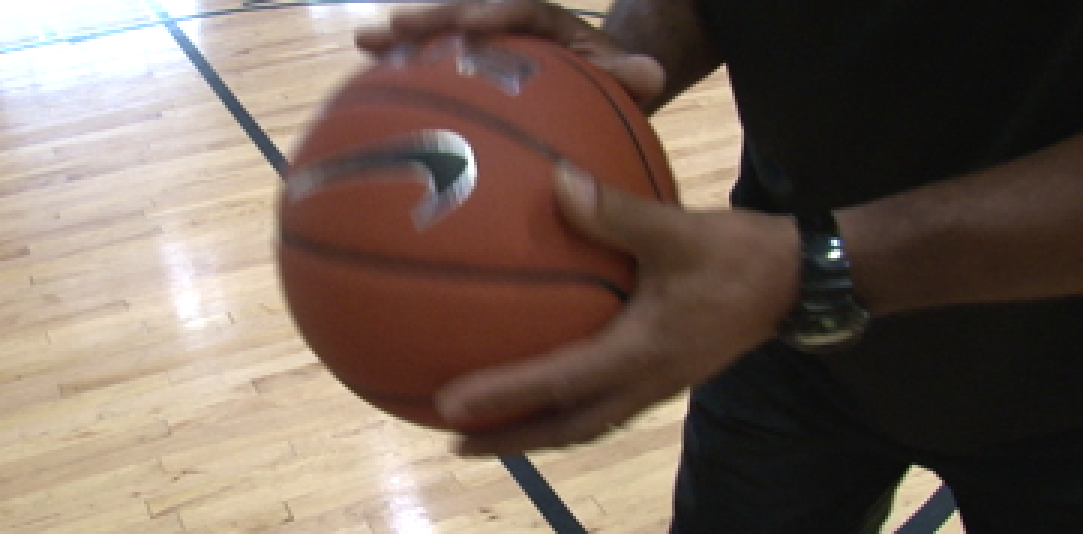}
  \subcaption{HR}
  \end{subfigure}
  \begin{subfigure}[b]{0.18\linewidth}
     \includegraphics[width=\linewidth]{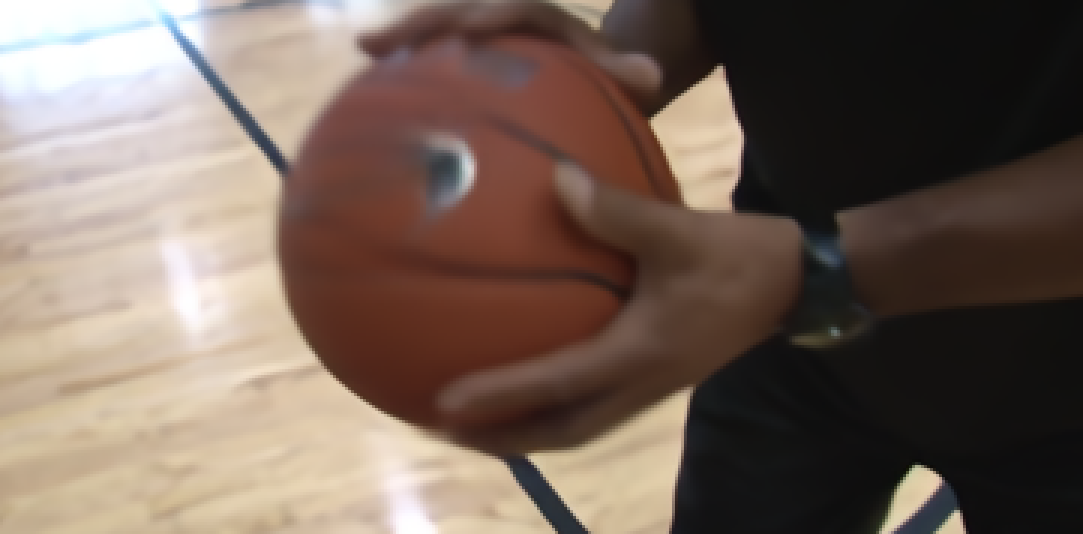}
     \subcaption{w/ DFI}
  \end{subfigure}
  \begin{subfigure}[b]{0.18\linewidth}
  \includegraphics[width=\linewidth]{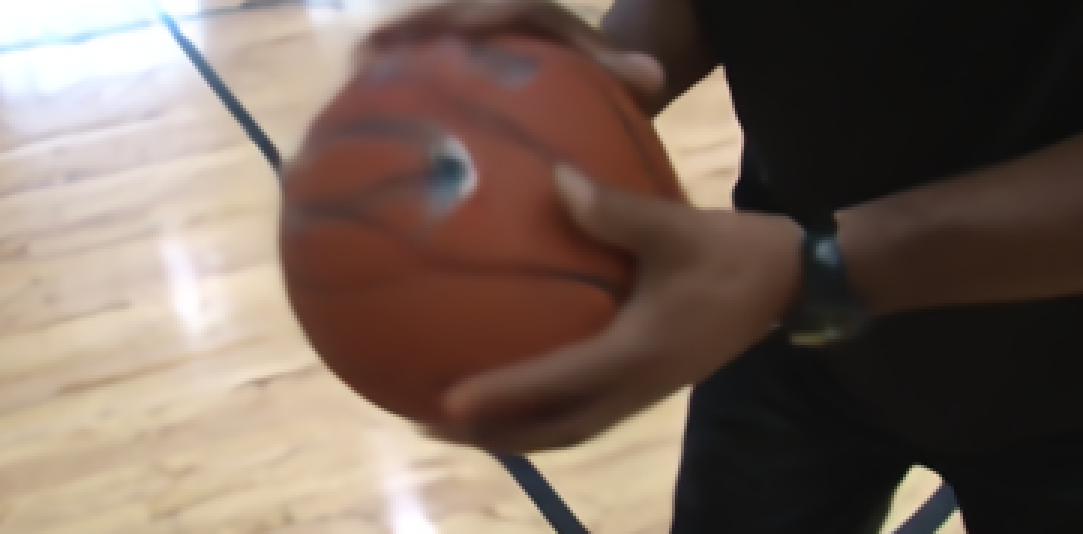}
  \subcaption{w/ DFI+ConvLSTM}
  \end{subfigure}
  \begin{subfigure}[b]{0.18\linewidth}
  \includegraphics[width=\linewidth]{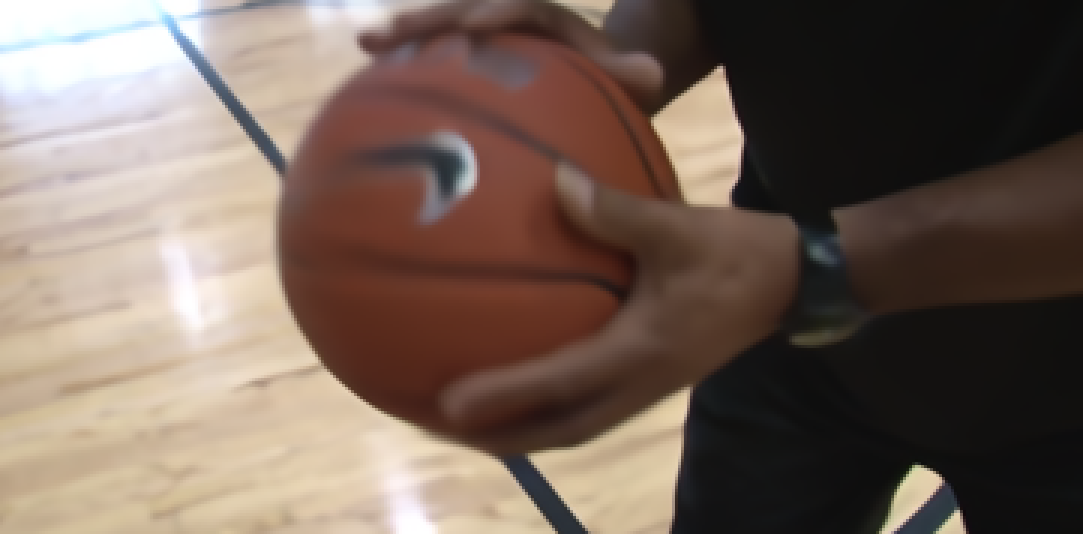}
  \subcaption{w/ DFI+DConvLSTM}
  \end{subfigure}
\end{center}
\vspace{-5mm}
   \caption{Ablation study on Deformable ConvLSTM (DConvLSTM). ConvLSTM will fail when meeting videos with fast motions. Embedded with state updating cells, the proposed DConvLSTM is more capable of leveraging global temporal contexts for reconstructing more accurate visual content even for fast motion videos.
   }
 \label{fig:ablation_dconvlstm}
\vspace{-4mm}
\end{figure*}

\begin{figure}[tb]
\captionsetup[subfigure]{labelformat=empty}
\begin{center}
  \begin{subfigure}[b]{0.32\linewidth}
     \includegraphics[width=\linewidth]{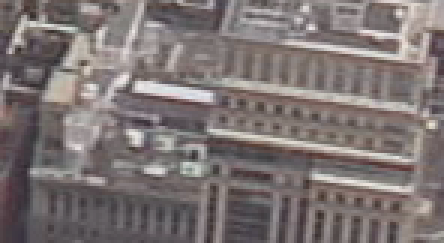}
     \subcaption{HR}
  \end{subfigure}
  \begin{subfigure}[b]{0.32\linewidth}
  \includegraphics[width=\linewidth]{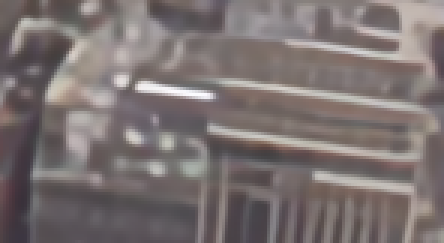}
  \subcaption{w/o bidirectional}
  \end{subfigure}
  \begin{subfigure}[b]{0.32\linewidth}
     \includegraphics[width=\linewidth]{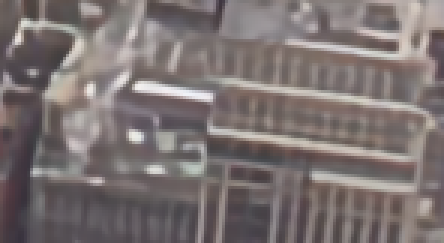}
     \subcaption{w/ bidirectional}
  \end{subfigure}
\end{center}
\vspace{-5mm}
   \caption{Ablation study on the bidirectional mechanism in DConvLSTM. Adding the bidirectional mechanism into DConvLSTM, the model can leverage both previous and future contexts, and therefore reconstructs more visually appealing frames with finer image details, especially for video frames at the first time step, which can not access any temporal information from other frames.}
 \label{fig:ablation_bidirectional}
\vspace{-4mm}
\end{figure}

\begin{description}[style=unboxed,leftmargin=0cm]
\item[Effectiveness of Deformable Feature Interpolation] To investigate the proposed deformable feature interpolation (DFI) module, we introduce two baselines: (a) and (b), where the model (a) only uses convolutions to blend LR features without deformable sampling functions as in model (b). In addition, neither (a) or (b) has ConvLSTM or DConvLSTM. From Table \ref{tab:ablation}, we find that (b) outperforms (a) by 0.16dB on Vid4 with slow motions and 0.73dB on Vimeo-Fast with fast motions in terms of PSNR. 
Figure~\ref{fig:ablation_interp} shows a visual comparison. We can see that (a) produces a face with severe motion blur, while the proposed deformable feature interpolation with exploiting local temporal contexts can effectively address the large motion issue and help the model (b) generate a frame with more clear face structures and details. The superiority of the proposed DFI module demonstrates that the learned offsets in the deformable sampling functions can effectively exploit local temporal contexts and successfully capture forward and backward motions even without any explicit supervision.
\end{description}


\begin{description}[style=unboxed,leftmargin=0cm]
\item[Effectiveness of Deformable ConvLSTM] 
To validate the effect of the proposed Deformable ConvLSTM (DConvLSTM), we compare four different models: (b), (c), (d), and (e), where (c) adds a vanilla ConvLSTM structure into (b), (d) utilizes the proposed DConvLSTM, and (e) adopts a DConvLSTM in a bidirectional manner.  

From Table \ref{tab:ablation}, we can see that (c) outperforms (b) on Vid4 with slow motion videos while it is worse than (b) on Vimeo-Fast with fast motion sequences. The results validate that vanilla ConvLSTM can leverage useful global temporal contexts for slow motion videos, but cannot handle large motions in videos. Moreover, we observe that (d) is significantly better than both (b) and (c), which demonstrates that our DConvLSTM can successfully learn the temporal alignment between previous states and the current feature map. Therefore, it can better exploit global contexts for reconstructing visually pleasing frames with more details. Visual results in Figure~\ref{fig:ablation_dconvlstm} further support our findings. 

In addition, we compare (e) and (d) in Table \ref{tab:ablation} and Figure~\ref{fig:ablation_bidirectional} to verify the bidirectional mechanism in DConvLSTM. From Table \ref{tab:ablation}, we can see that (e) can further improve STVSR performance over (d) on both slow motion and fast motion testing sets. The visual results in Figure~\ref{fig:ablation_bidirectional} further shows that our full model with a bidirectional mechanism can restore more visual details by making full use of global temporal information for all input video frames.

\end{description}

\section{Conclusion}
In this paper, we propose a one-stage framework for space-time video super-resolution to directly reconstruct high-resolution and high frame rate videos without synthesizing intermediate low-resolution frames. To achieve this, we introduce a deformable feature interpolation network for feature-level temporal interpolation. Furthermore, we propose a deformable ConvLSTM for aggregating temporal information and handling motions. With such a one-stage design, our network can well explore intra-relatedness between temporal interpolation and spatial super-resolution in the task. It enforces our model to adaptively learn to leverage useful local and global temporal contexts for alleviating large motion issues. 
Extensive experiments show that our one-stage framework is more effective yet efficient than existing two-stage networks, and the proposed feature temporal interpolation network and deformable ConvLSTM are capable of handling very challenging fast motion videos.


\section*{Acknowledgements}
\noindent The work was partly supported by NSF 1741472, 1813709, and 1909912.
This article solely reflects the opinions and conclusions of its authors but not the funding agents.

{\small
\bibliographystyle{ieee_fullname}
\bibliography{egbib}
}

\onecolumn
\begin{appendices}
\section*{Network Architecture}
\begin{figure}[htbp]
    \centering
    \includegraphics[width=0.4\columnwidth]{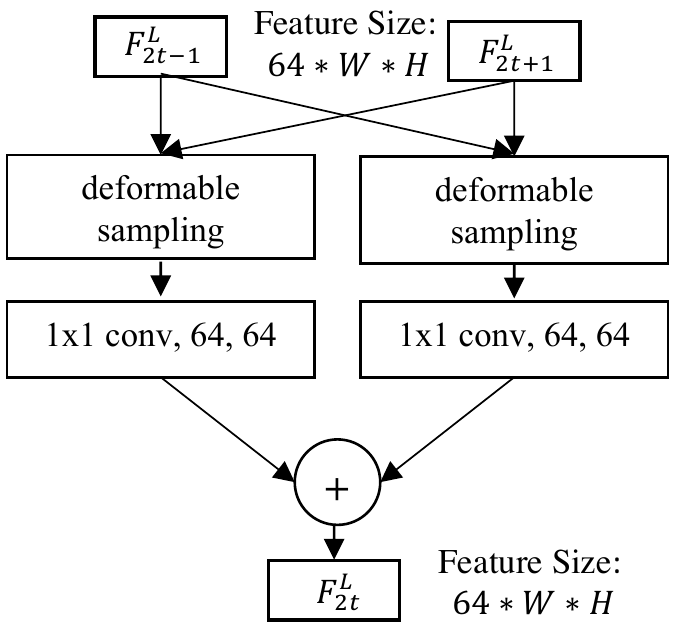}
    \caption{Feature temporal interpolation for intermediate LR frames. It will predict an intermediate LR frame feature map $F_{2t}^{L}$ from two neighboring feature maps: $F_{2t-1}^{L}$ and $F_{2t+1}^{L}$, where $t$ = 1, 2, ..., $n$. Note that the deformable sampling module on the left samples features from $F_{2t-1}^{L}$ with generated sampling parameters from both $F_{2t-1}^{L}$ and $F_{2t+1}^{L}$; on the contrary, the deformable sampling module on the right samples features from $F_{2t+1}^{L}$.} 
    \label{fig:finterp}
\vspace{-5mm}
\end{figure}

\begin{figure*}[htbp]
    \centering
    \includegraphics{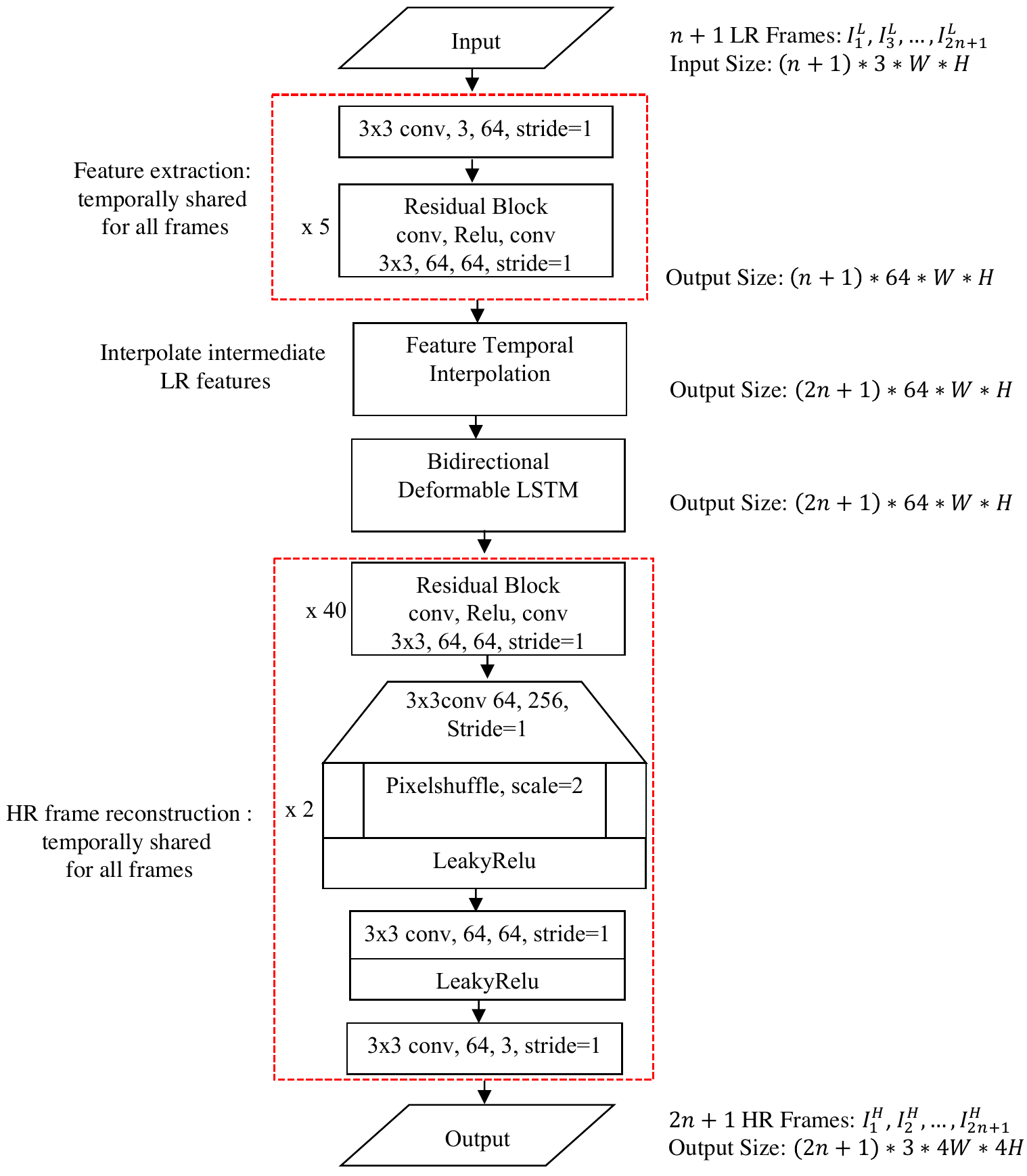}
    \caption{Flowchart of the proposed one-stage STVSR framework. The feature extraction and HR frame reconstruction networks are temporally shared for all frames, in which different frames are processed independently.}
    \label{fig:net}
\vspace{-5mm}
\end{figure*}

We further illustrate the feature temporal interpolation network in Figure~\ref{fig:finterp} and the proposed STVSR framework in Figure~\ref{fig:net} to help readers better understand the overall structure of our proposed network. 

To make our paper be concise and easy to follow, we use a simple version of deformable sampling to introduce the proposed feature temporal interpolation and deformable ConvLSTM. However, in our implementation, as stated in Section \textcolor{red}{3.4} of the paper, we adopt a Pyramid, Cascading and Deformable (PCD) structure\footnote{The official PyTorch implementation of the PCD can be found in \href{https://github.com/xinntao/EDVR}{https://github.com/xinntao/EDVR}.} as in~\cite{wang2019edvr} to implement the deformable sampling, which can exploit multi-scale contexts with a feature pyramid. 

\end{appendices}

\end{document}